\documentclass{article}

\PassOptionsToPackage{numbers, compress}{natbib}
\usepackage[final]{nips_2018} 
\usepackage{graphicx}
\usepackage{float}

\usepackage{xcolor}

\usepackage{algorithmic}
\usepackage{algorithm}

\usepackage[utf8]{inputenc} 
\usepackage[T1]{fontenc}    
\usepackage{hyperref}       
\usepackage{url}            
\usepackage{booktabs}       
\usepackage{amsfonts}       
\usepackage{nicefrac}       
\usepackage{microtype}      

\usepackage{amsmath, bm}
\usepackage{amssymb}

\title{Practical Deep Reinforcement Learning Approach for Stock Trading}

\author{
Xiao-Yang Liu$^{1,*}$, Zhuoran Xiong$^{1,*}$, 
       Shan Zhong$^{1}$, 
       Hongyang (Bruce) Yang$^{2}$,
       and Anwar Walid$^3$\\
       $^1$Electrical Engineering, Columbia University,\\
       $^2$Department of Statistics, Columbia University,\\
       $^3$Mathematics of Systems Research Department, Nokia-Bell Labs\\
       Emails: \{ZX2214, XL2427, SZ2495, HY2500\}@columbia.edu, \\  anwar.walid@nokia-bell-labs.com       
       }

\begin{document}

\maketitle

\begin{abstract}


 Stock trading strategy plays a crucial role in investment companies. However, it is challenging to obtain optimal strategy in the complex and dynamic stock market. We explore the potential of deep reinforcement learning to optimize stock trading strategy and thus maximize investment return. $30$ stocks are selected as our trading stocks and their daily prices are used as the training and trading market environment. We train a deep reinforcement learning agent and obtain an adaptive trading strategy. The agent's performance is evaluated and compared with Dow Jones Industrial Average and the traditional min-variance portfolio allocation strategy. The proposed deep reinforcement learning approach is shown to outperform the two baselines in terms of both the Sharpe ratio and cumulative returns.
  
\end{abstract}

\section{Introduction}
\label{introduction}

Profitable stock trading strategy is vital to investment companies. It is applied to optimize allocation of capital and thus maximize performance, such as expected return. Return maximization is based on estimates of stocks' potential return and risk. However, it is challenging for analysts to take all relavant factors into consideration in complex stock market \cite{fuzzy, online, intelligence}.

One traditional approach is performed in two steps as described in \cite{selection}. First, the expected returns of the stocks and the covariance matrix of the stock prices are computed. The best portfolio allocation is then found by either maximizing the return for a fixed risk of the portfolio or minimizing the risk for a range of returns. The best trading strategy is then extracted by following the best portfolio allocation. This approach, however, can be very complicated to implement if the manager wants to revise the decisions made at each time step and take, for example, transaction cost into consideration. Another approach to solve the stock trading problem is to model it as a Markov Decision Process (MDP) and use dynamic programming to solve for the optimum strategy. However, the scalability of this model is limited due to the large state spaces when dealing with the stock market \cite{DP,Testing,Optimum, Enhance}.

Motivated by the above challenges, we explore a deep reinforcement learning algorithm, namely Deep Deterministic Policy Gradient (DDPG) \cite{DDPG}, to find the best trading strategy in the complex and dynamic stock market. This algorithm consists of three key components: (i) actor-critic framework \cite{actor} that models large state and action spaces; (ii) target network that stabilizes the training process \cite{DQN}; (iii) experience replay that removes the correlation between samples and increases the usage of data. The efficiency of DDPG algorithm is demonstrated by achieving higher return than the traditional min-variance portfolio allocation method and the Dow Jones Industrial Average \footnote{The Dow Jones Industrial Average is a stock market index that shows how 30 large, publicly owned companies based in the United States have traded during a standard trading session in the stock market.} (DJIA).


This paper is organized as follows. Section 2 contains statement of our stock trading problem. In Section 3, we drive and specify the main DDPG algorithm. Section 4 describes our data preprocessing and  experimental setup, and presents the performance of DDPG algorithm. Section 5 gives our conclusions.

\section{Problem Statement}
\label{reinforcement}
We model the stock trading process as a Markov Decision Process (MDP). We then formulate our trading goal as a maximization problem. 

\subsection{Problem Formulation for Stock Trading}

\begin{figure}[t]
\centering
\includegraphics[scale = 0.4]{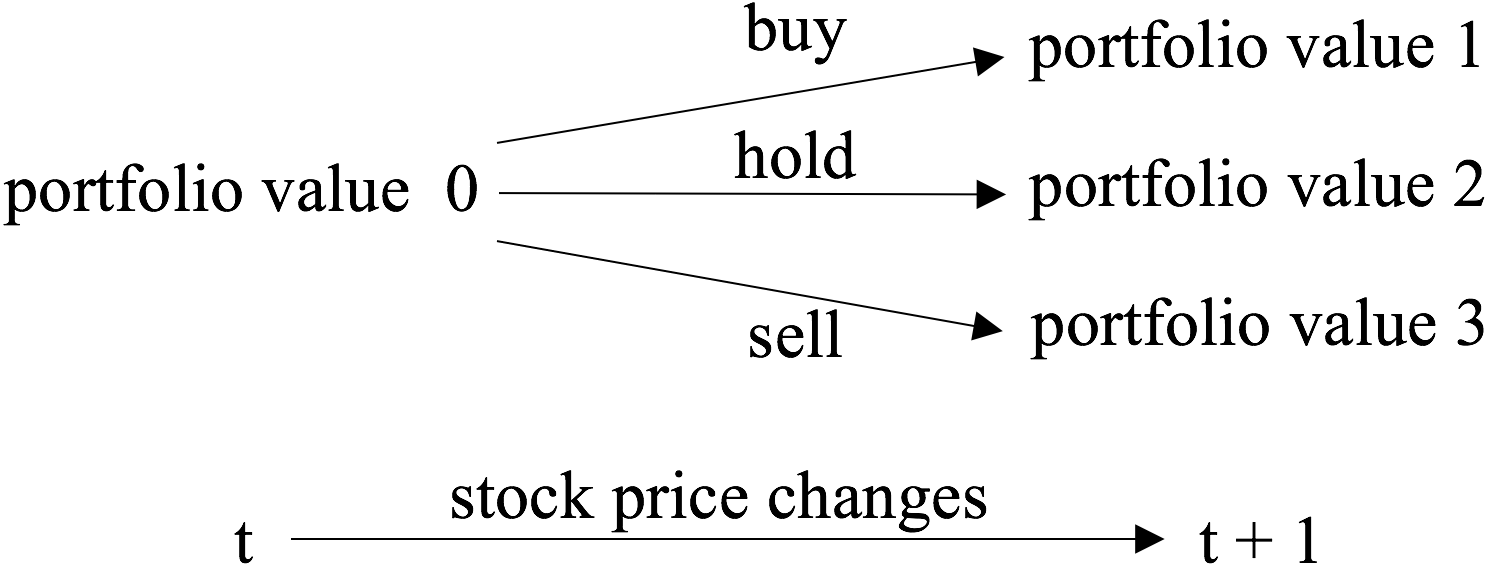}
\caption{One starting portfolio value with three actions leading to three possible portfolio values where actions have probabilities that sum up to one. Note that "hold" can lead to different portfolio values if the stock prices change.}
\label{iteration0}

\end{figure}

Considering the stochastic and interactive nature of the trading market, we model the stock trading process as a Markov Decision Process (MDP) as shown in Fig. \ref{iteration0}, which is specified as follows:
\begin{itemize}
\item State $s= [p,h,b]$: a set that includes the information of the prices of stocks $p\in \mathbb{R}_+^D$, the amount of holdings of stocks $h\in \mathbb{Z}_+^D$, and the remaining balance $b\in \mathbb{R}_+$, where $D$ is the number of stocks that we consider in the market and $\mathbb{Z}_+$ denotes non-negative integer numbers.
\item Action $a$: a set of actions on all $D$ stocks. The available actions of each stock include selling, buying, and holding, which result in decreasing, increasing, and no change of the holdings $h$, respectively. 
\item Reward $r(s,a,s')$: 
the change of the portfolio value when action $a$ is taken at state $s$ and arriving at the new state $s'$. The portfolio value is the sum of the equities in all held stocks $p^T h$ and balance $b$.
\item Policy $\pi(s)$: the trading strategy of stocks at state $s$. It is essentially the probability distribution of $a$ at state $s$.
\item Action-value function $Q_\pi(s,a)$: the expected reward achieved by action $a$ at state $s$ following policy $\pi$.
\end{itemize}
The dynamics of the stock market is described as follows. We use subscript to denote time $t$, and the available actions on stock $d$ are
\begin{itemize}
    \item Selling: $k$ ($k\in [1,h[d]]$, where $d = 1,...,D$) shares can be sold from the current holdings, where $k$ must be an integer. In this case, $h_{t+1}=h_t-k$.
    \item Holding: $k=0$ and it leads to no change in $h_t$.
    \item Buying: $k$ shares can be bought and it leads to $h_{t+1}=h_t+k$. In this case $a_t[d]=-k$ is a negative integer.
\end{itemize}
It should be noted that all bought stocks should not result in a negative balance on the portfolio value. That is, without loss of generality we assume that selling orders are made on the first $d_1$ stocks and the buying orders are made on the last $d_2$ ones, and that $a_t$ should satisfy $p_t[1:d_1]^T a_t[1:d_1]+b_t+ p_t[D-d_2:D]^T a_t[D-d_2:D]\ge 0$. The remaining balance is updated as $b_{t+1}=b_t+p_t^T a_t$.
Fig. \ref{iteration0} illustrates this process. As defined above, the portfolio value consists of the balance and sum of the equities in all held stocks. 
At time $t$, an action is taken, and based on the executed action and the updates of stock prices, the portfolio values change from "portfolio value 0" to "portfolio value 1", "portfolio value 2", or "portfolio value 3" at time $(t+1)$. 

Before being exposed to the environment, $p_0$ is set to the stock prices at time $0$ and $b_0$ is the initial fund available for trading. The $h$ and $Q_{\pi}(s,a)$ are initialized as 0, and $\pi(s)$ is uniformly distributed among all actions for any state. Then, $Q_{\pi}(s_{t},a_{t})$ is learned through interacting with the external environment.


According to Bellman Equation, the expected reward of taking action $a_t$ is calculated by taking the expectation of the rewards $r(s_t,a_t,s_{t+1})$, plus the expected reward in the next state $s_{t+1}$. Based on the assumption that the returns are discounted by a factor of $\gamma$, we have
\begin{equation}
    \label{decom}
    Q_\pi(s_{t},a_t) = \mathbb{E}_{s_{t+1}}[r(s_t,a_t,s_{t+1}) + \gamma \mathbb{E}_{a_{t+1} \sim \pi(s_{t+1})}[ Q_\pi (s_{t+1},a_{t+1})]].
\end{equation}

\subsection{Trading Goal as Return Maximization}

The goal is to design a trading strategy that maximizes the investment return at a target time $t_f$ in the future, i.e., $p_{t_f}^T h_t+b_{t_f}$, which is also equivalent to $\sum_{t=1}^{t_f-1}r(s_t,a_t,s_{t+1})$. Due to the Markov property of the model, the problem can be boiled down to optimizing the policy that maximizes the function $Q_\pi (s_t,a_t)$. This problem is very hard because the action-value function is unknown to the policy maker and has to be learned via interacting with the environment. Hence in this paper, we employ the deep reinforcement learning approach to solve this problem.


\section{A Deep Reinforcement Learning Approach}
We employ a DDPG algorithm to maximize the investment return. DDPG is an improved version of Deterministic Policy Gradient (DPG) algorithm \cite{DPG}. DPG combines the frameworks of both Q-learning \cite{Sutton} and policy gradient \cite{Policy Gradient}. Compared with DPG, DDPG uses neural networks as function approximator. The DDPG algorithm in this section is specified for the MDP model of the stock trading market. 

The Q-learning is essentially a method to learn the environment. Instead of using the expectation of $Q(s_{t+1},a_{t+1})$ to update $Q(s_{t},a_{t})$, Q-learning uses greedy action $a_{t+1}$ that maximizes $Q(s_{t+1},a_{t+1})$ for state $s_{t+1}$, i.e.,
\begin{equation}
    \label{Q*}
     Q_\pi(s_t,a_t) = \mathbb{E}_{s_{t+1}}[r(s_t,a_t,s_{t+1}) + \gamma  \max_{a_{t+1}}Q (s_{t+1},a_{t+1})].
\end{equation}


\begin{figure}[t]
\centering
\includegraphics[scale = 0.5]{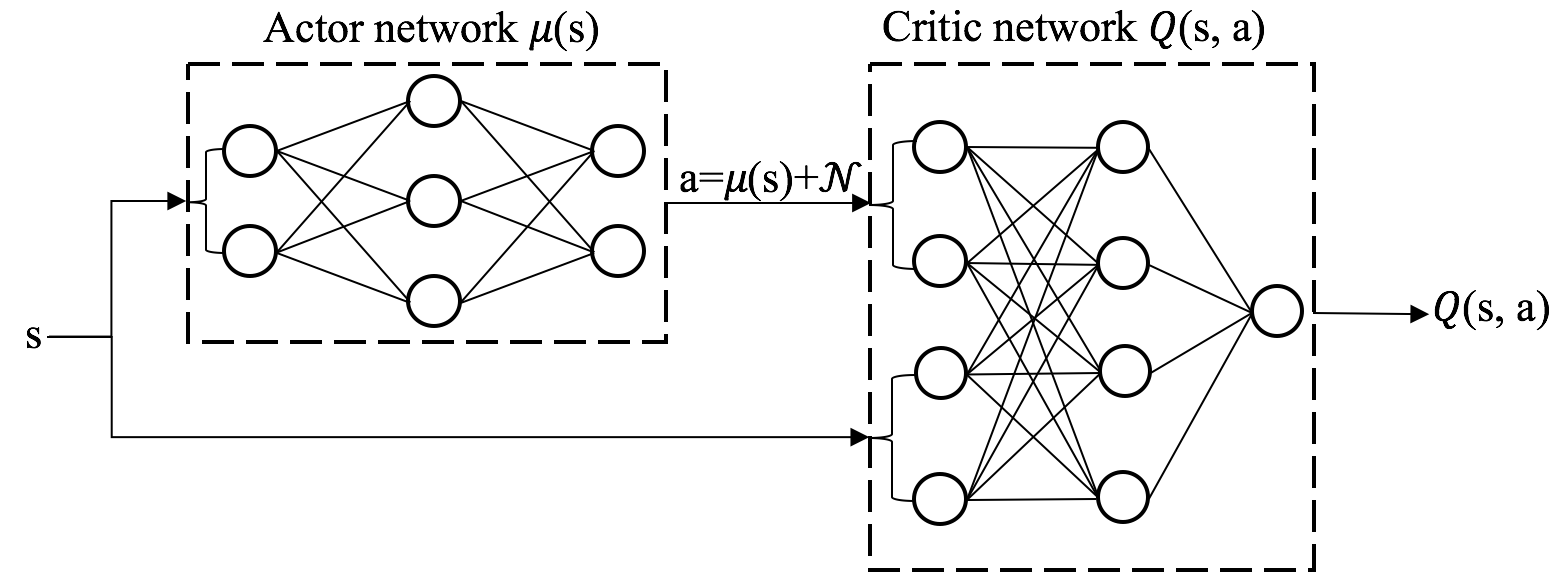}
\caption{Learning network achitecture.}
\label{nn}

\end{figure}

With Deep Q-network (DQN), which adopts neural networks to perform function approximation, the states are encoded in value function. The DQN approach, however, is intractable for this problem due to the large size of the action spaces. Since the feasible trading actions for each stock is in a discrete set, and considering the number of total stocks, the sizes of action spaces grow exponentially, leading to the "curse of dimensionality" \cite{Management}. Hence, the DDPG algorithm is proposed to deterministically map states to actions to address this issue.

As shown in Fig. \ref{nn}, DDPG maintains an actor network and a critic network. The actor network $\mu(s| \theta^\mu)$ maps states to actions where $\theta^{\mu}$ is the set of actor network parameters, and the critic network $Q(s,a|\theta^Q)$ outputs the value of action under that state, where $\theta^{Q}$ is the set of critic network parameters. To explore better actions, a noise is added to the output of the actor network, which is sampled from a random process $\mathcal{N}$.

\begin{algorithm}[t]
	\caption{DDPG algorithm}
	\begin{algorithmic}[1]
		\STATE Randomly initialize critic network $Q(s,a|\theta^Q)$ and actor $\mu(s|\theta^\mu)$ with random weight $\theta^Q$ and $\theta^\mu$;
        \STATE Initialize target network $Q'$ and $\mu' $ with weights $\theta^{Q'} \leftarrow \theta^{Q}$, $\theta^{\mu'} \leftarrow \theta^{\mu}$;
        \STATE Initialize replay buffer $R$;
        \FOR {episode= 1, $M$}
            \STATE Initialize a random process $\mathcal{N}$ for action exploration;
            \STATE Receive initial observation state $s_1$;
            \FOR{t = 1, $T$}
                \STATE {Select action $a_t = \mu(s_t|\theta^\mu) + \mathcal{N}_t$ according to the current policy and exploration noise;}
                \STATE {Execute action $a_t$ and observe reward $r_t$ and state $s_{t+1}$};
                \STATE {Store transition ($s_t$, $a_t$, $r_t$, $s_{t+1})$ in $R$;}
                \STATE {Sample a random minibatch of $N$ transitions ($s_i$ , $a_i$ , $r_i$ , $s_{i +1}$) from $R$;}
                \STATE {Set $y_i = r_i+\gamma Q' (s_{t+1}, \mu' (s_{i+1}|\theta^{\mu'}|\theta^{Q'}))$;}
                \STATE {Update critic by minimizing the loss: $L = \frac{1}{N}\sum_i(y_i -Q(s_i,a_i|\theta^Q))^2$;}
                \STATE {Update the actor policy by using the sampled policy gradient:
                $$
                    \nabla_{\theta^\mu} J\approx \frac{1}{N}\sum_i \nabla_a Q(s,a|\theta^Q)|_{s = s_i,a = \mu(s_i)} \nabla_{\theta^\mu} \mu(s|\theta^\mu)|_{s_i};
                $$}
                \STATE Update the target networks:
                $$\theta^{Q'} \leftarrow \tau \theta^Q + (1-\tau)\theta^{Q'},
                $$ $$
                \theta^{\mu'} \leftarrow \tau \theta^\mu + (1-\tau)\theta^{\mu'}.
                $$
            \ENDFOR
        \ENDFOR
	\end{algorithmic}
	\vspace{-2pt}
	\label{algo}
\end{algorithm}

Similar to DQN, DDPG uses an experience replay buffer $R$ to store transitions and update the model, and can effectively reduce the correlation between experience samples. Target actor network $Q'$ and $\mu'$ are created by copying the actor and critic networks respectively, so that they provide consistent temporal difference backups. Both networks are updated iteratively.
At each time, the DDPG agent takes an action $a_t$ on $s_t$, and then receives a reward based on $s_{t+1}$. The transition $(s_t,a_t,s_{t+1},r_t)$ is then stored in replay buffer $R$. The $N$ sample transitions are drawn from $R$ and $y_i = r_i+\gamma Q' (s_{i+1}, \mu' (s_{i+1}|\theta^{\mu'},\theta^{Q'})), i=1,\cdots,N,$ is calculated. The critic network is then updated by minimizing the expected difference $L(\theta^Q)$ between outputs of the target critic network $Q'$ and the critic network $Q$, i.e,
\begin{equation}
    L(\theta^Q) = \mathbb{E}_{s_t,a_t,r_t,s_{t+1} \sim \text{buffer}}[(r_t+ \gamma Q'(s_{t+1},\mu (s_{t+1}|\theta^\mu)|\theta^{Q'}) - Q(s_t,a_t|\theta^Q))^2].
\end{equation} 
The parameters $\theta^\mu$ of the actor network are then as follows:
\begin{align}
    \nabla_{\theta^\mu} J &\approx \mathbb{E}_{s_t,a_t,r_t,s_{t+1}\sim \text{buffer}}
    [\nabla_{\theta^\mu} Q(s_t,\mu(s_t|\theta^\mu)|\theta^Q)] \\
    & = \mathbb{E}_{s_t,a_t,r_t,s_{t+1}\sim \text{buffer}}
    [\nabla_{a} Q(s_t,\mu(s_t)|\theta^Q) \nabla_{\theta^\mu} \mu(s_t|\theta^\mu)].
\end{align}
After the critic network and the actor network are updated by the transitions from the experience buffer, the target actor network and the target critic network are updated as follows:
\begin{align}
    \theta^{Q'} &\leftarrow \tau \theta^Q + (1-\tau)\theta^{Q'},
    \\
    \theta^{\mu'} &\leftarrow \tau \theta^\mu + (1-\tau)\theta^{\mu'},
\end{align}

where $\tau$ denotes learning rate. The detailed algorithm is summerized in Algorithm \ref{algo}.

\section{Performance Evaluations}
We evaluate the performance of the DDPG algorithm in Alg. \ref{algo}. The result demonstrates that the proposed method with the DDPG agent achieves higher return than the Dow Jones Industrial Average and the traditional min-variance portfolio allocation strategy \cite{Codes, ML}.

\begin{figure}[ht]
\centering
\includegraphics[scale=0.5]{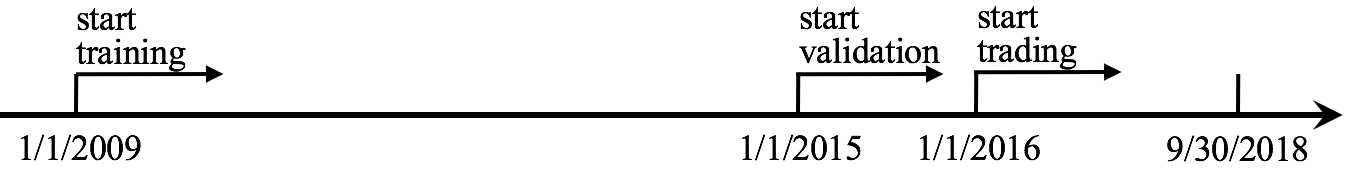}
\caption{Data spliting.}
\label{data}
\end{figure}

\subsection{Data Preprocessing}
We track and select Dow Jones $30$ stocks of 1/1/2016 as our trading stocks, and use historical daily prices from 01/01/2009 to 09/30/2018 to train the agent and test the performance. The dataset is downloaded from Compustat database accessed through Wharton Research Data Services (WRDS) \cite{Data}. 

Our experiment consists of three stages, namely training, validation and trading. In the training stage, Alg. 1 generates a well-trained trading agent. The validation stage is then carried out for key parameters adjustment such as learning rate, number of episodes, etc. Finally in the trading stage, we evaluate the profitability of the proposed scheme. 
The whole dataset is split into three parts for these purposes, as shown in Fig. \ref{data}. Data from 01/01/2009 to 12/31/2014 are used for training, and the data from 01/01/2015 to 01/01/2016 are used for validation. We train our agent on both training and validation data to make full use of available data. Finally, we test our agent's performance on trading data, which is from 01/01/2016 to 09/30/2018. To better exploit the trading data, we continue training our agent while in the trading stage as this will improve the agent to better adapt the market dynamics.

\subsection{Experimental Setting and Results of Stock Trading}

We build the environment by setting 30 stocks data as a vector of daily stock prices over which the DDPG agent is trained. To update the learning rate and number of episodes, the agent is validated on validation data. Finally, we run our agent on trading data and compare performance with Dow Jones Industrial Average (DJIA) and the min-variance portfolio allocation strategy. 


Four metrics are used to evaluate our results: final portfolio value, annualized return, annualized standard error and the Sharpe ratio. Final portfolio value reflects portfolio value at the end of trading stage. Annualized return indicates the direct return of the portfolio per year. Annualized standard error shows the robustness of our model. The Sharpe ratio combines the return and risk together to give such evaluation \cite{Sharpe}.

\begin{figure}[ht]
\centering
\includegraphics[scale=0.7]{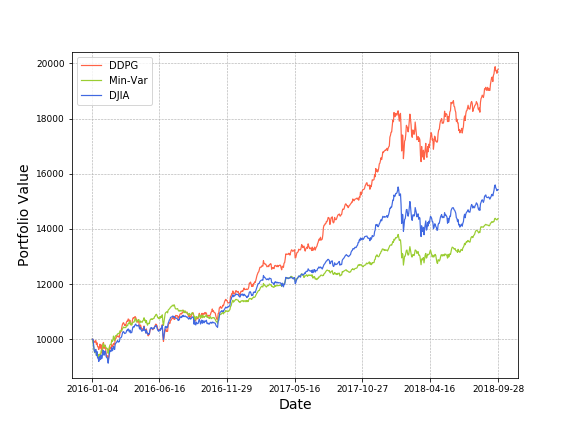}
\caption{Portfolio value curves of our DDPG scheme, the min-variance portfolio allocation strategy, and the Dow Jones Industrial Average. (Initial portfolio value \$$10, 000$).}
\label{comparing}
\end{figure}

In Fig. \ref{comparing}, we can see that the DDPG strategy significantly outperforms Dow Jones Industrial Average and the min-variance portfolio allocation. As can be seen from Table 1, the DDPG strategy achieves annualized return $22.24\%$, which is much higher than Dow Jones Industrial Average's $16.40\%$ and min-variance portfolio allocation's $15.93\%$. The sharpe ratio of the DDPG strategy is also much higher, indicating that the DDPG strategy beats both Dow Jones Industrial Average and min-variance portfolio allocation in balancing risk and return. Therefore, the result demonstrates that the proposed DDPG strategy can effectively develop a trading strategy that outperforms the benchmark Dow Jones Industrial Average and the traditional min-variance portfolio allocation method.


\begin{table}[ht]
    \centering
    \caption{Trading Performance.}
    \begin{tabular}{|c|c|c|c|}
    \hline
        &\textbf{DDPG} (\textbf{ours}) & Min-Variance &  DJIA\\
         \hline
         Initial Portfolio Value & $\bm{10, 000}$  & $10,000$ & $10,000$\\
        \hline
        Final Portfolio Value & $\bm{19,791}$  & $14,369$ & $15,428$\\
        \hline
        Annualized Return & $\bm{25.87\%}$   & $15.93\%$ & $16.40\%$ \\
         \hline
       Annualized Std. Error  & $\bm{13.62\%}$ & $9.97\%$ & $11.70\%$ \\
          \hline
      Sharpe Ratio & $\bm{1.79}$ & $1.45$ & $1.27$\\
    \hline
    \end{tabular}
\end{table}

\small

\section{Conclusion}
In this paper, we have explored the potential of training Deep Deterministic Policy Gradient (DDPG) agent to learn stock trading strategy. Results show that our trained agent outperforms the Dow Jones Industrial Average and min-variance portfolio allocation method in accumulated return. The comparison on Sharpe ratios indicates that our method is more robust than the others in balancing risk and return. 

Future work will be interesting to explore more sophisticated model \cite{RNN}, deal with larger scale data \cite{Large}, observe intelligent behaviors \cite{ITS2018}, and incorporate prediction schemes \cite{Time_Series}.

\end{document}